\documentclass[runningheads]{llncs}
\usepackage{graphicx}
\usepackage{wrapfig}
\usepackage{subfigure}
\usepackage{amsmath}
\usepackage{multirow}

\begin{document}
\title{Robustness of Neural Networks to\\Parameter Quantization}
%
%
\author{Abhishek Murthy\inst{1} \and
Himel Das\inst{2} \and
Md. Ariful Islam\inst{2}}
\authorrunning{A. Murthy et al.}
%
\institute{Signify Research North Americas\footnote{Research was performed
with other authors at Texas Tech University.}, Cambridge, MA,
\email{amurthy.sunysb@gmail.com}
\and
Texas Tech University, Lubbock, TX\\
\email{\{himel.das, ariful.islam\}@ttu.edu}}
\maketitle              
\begin{abstract}

         
Quantization, a commonly used technique to reduce the memory
footprint of a neural network for edge computing, entails
reducing the precision of the floating-point representation
used for the parameters of the network.  The impact of such
rounding-off errors on the overall performance of the neural
network is estimated using testing, which is not exhaustive
and thus cannot be used to guarantee the safety of the model.
We present a framework based on Satisfiability Modulo Theory
(SMT) solvers to quantify the robustness of neural networks
to parameter perturbation.  To this end, we introduce notions
of local and global robustness that capture the deviation in
the confidence of class assignments due to parameter quantization.
The robustness notions are then cast as instances of SMT problems
and solved automatically using solvers, such as dReal.  We
demonstrate our framework on two simple Multi-Layer Perceptrons
(MLP) that perform binary classification on a two-dimensional
input.  In addition to quantifying the robustness, we also show
that Rectified Linear Unit activation results in higher robustness
than linear activations for our MLPs.
\end{abstract}
\section{Introduction}
\label{sec:intro}
Neural networks entail interconnected computational nodes
that transform \\ weighted combinations of their inputs using
nonlinear functions.  The interconnections lead to
compositional behavior at the network-level, which enables
neural networks to approximate highly nonlinear functions as
their responses.  The advent of the Backpropagation 
algorithm~\cite{bprop:hinton}, the availability of large
datasets~\cite{imagenet}, and optimized hardware~\cite{nvidia}
has led to widespread success in supervised and unsupervised
learning.

\begin{figure}[!ht]
\centering
\includegraphics[width=1\linewidth]{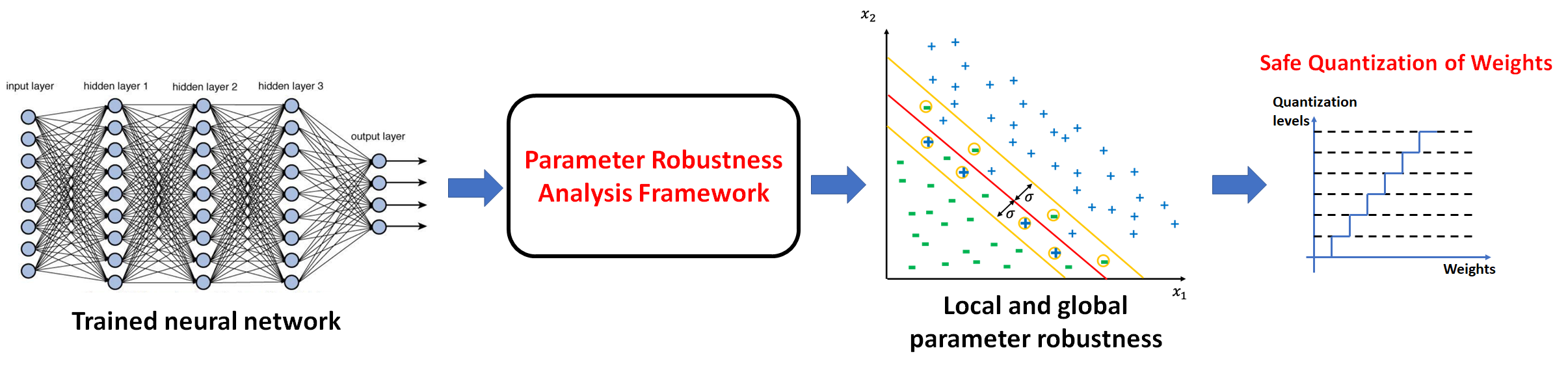}
\vspace{-2ex}
\caption{Robustness analysis of neural network enables safe
         parameter quantization.}
\label{fig:overview}
\vspace*{-4ex}
\end{figure}

Supervised learning of a neural network is the process
of optimizing the network's parameters using reference data.
Supervised learning can be used to i) learn classifiers,
which can label an input into one of finitely many classes
and ii) learn the more general class of regressors, which
capture relationships across continuous domains.  Learning
a model, also known as training, involves formulating a
loss function that quantifies the performance of the model
as a function of the parameters, and then minimizing the
function using numerical techniques over the reference data,
also known as training data.  Backpropagation is
the most popular class of numerical techniques used to
optimize the parameters of the modern neural networks.
Unsupervised learning, on the other hand, entails learning
patterns and underlying distributions in unlabelled data.


Large networks contain millions 
of parameters and are trained using Graphics Processing Units
(GPUs).  Deploying trained neural networks in real-world production systems
entails fetching
the input from the user/client-device and then passing it through
the neural network, also known as the forward pass, and obtaining
the output, which could be a class-label or a regressed value,
\emph{in real time}.  Web services, which perform the
forward pass on the cloud can utilize the power of GPUs for
time-sensitive calculations.  The downside is that such applications
suffer from i) the latency of sending the input to the remote server
and waiting for the output of the neural network and ii) privacy
concerns of exposing potentially sensitive inputs on the network.

An alternative design involves performing the forward pass on the
client device (edge) by running the neural networks on it.  This
eliminates the network latencies and also avoids exposing the
user's inputs to the network.  Running neural networks on edge
devices, such as mobile phones, tablets and low-power devices like
wearables and Raspberry Pis present unique challenges.  Storing
the millions of parameters in floating-point representations
incurs significant memory costs and the computational power needed
for the forward pass may be prohobitive.  Executing complex neural
networks on the low computational power and memory available on
edge devices is a well-known challenge in the industry and thus
is an active area of interest.

In addition to dedicated hardware for low-power devices, the community
has evolved three main approaches to the problem of running neural
networks on resource-constrained edge devices.
\vspace{-1ex}
\begin{enumerate}
\item 
\emph{Quantization of Parameters}: The precision of the
floating-point representation used to store the network parameters
is reduced to lower the memory footprint of storing the network
in the memory~\cite{quantization}.
\item
\emph{Pruning}: The edges, represented by the weights, between nodes
that do not significantly influence the network's output are made 0
and thus removed from the network, resulting in a reduction in the
memory footprint~\cite{pruning}.
\item
\emph{Optimized Neural-Network Architectures}: The network
architecture is designed to reduce the floating-point operations,
thereby reducing the running time of the forward pass, see
\cite{squeezenet} for an example.
\end{enumerate}

These techniques have emerged using empirical benchmarking and
have found limited success in the community.  Today, there
are a handful of applications that deploy neural networks on edge
devices.  The main reason behind this lack of widespread success
is the unpredictability of the aforementioned techniques in preserving
the performance of the network after training.  Specifically, the
state of the art on estimating the impact of pruning and quantization
on the network's accuracy is limited to testing on a finite number
of test cases.  

\emph{In this paper, we introduce a framework to quantify the
robustness of neural networks to parameter quantization, thereby
automating the process of bounding the change in performance of
the neural network.}  We introduce notions of local and global
robustness of networks to parameter changes.  Given a bounded
perturbation in the parameter vector, local robustness
measures the maximal change in the confidence of class assignment for an input.
Global robustness extends this notion to the entire input-space.
We cast these notions into instances of SMT problems and solve
them automatically using solvers, such as dReal~\cite{dreal}.  See
Fig.~\ref{fig:overview} for an overview.

Robustness of neural networks has been an active area of research,
but most of the authors have focused on input perturbations, rather
than parameter changes.  Our framework is focused on parameter
perturbations.  
In summary, the main contributions of our paper are as follows.
\begin{itemize}
\item 
An automated framework is presented for bounding the deviation in
the performance of neural networks due to parameter quantization.
The framework enables the implementation of deep-learning-based
applications on edge devices, like mobile phones, tablets and
other embedded environments.

\item
We present two use-cases to demonstrate our framework: the parameters
of two small MLPs that perform binary classification are perturbed
and the robustness is analyzed using our approach.

\item
In addition to estimating parameter robustness, we also show
that ReLU activations are more robust than linear
activations for our MLPs.
\end{itemize}

The rest of the paper is organized as follows.
Section~\ref{sec:background} presents background on neural networks
and SMT solvers.  Section~\ref{sec:param_robust} introduces the theory
of local and global robustness to parameter perturbations and 
Section~\ref{sec:ver} details the
corresponding SMT problem formulations.
Section~\ref{sec:case_studies} presents
the case studies and their corresponding trained neural networks.
Section~\ref{sec:results} presents robustness analysis on the
neural networks.  Section~\ref{sec:related} reviews related work
and Section~\ref{sec:conclusions} presents our conclusions and
the directions for future work.
\vspace{4ex}

\vspace{6ex}
\section{Background}
\label{sec:background}
\begin{wrapfigure}{!h}{0.5\textwidth}
\vspace{-6ex}
\centering
\includegraphics[width=0.5\textwidth]{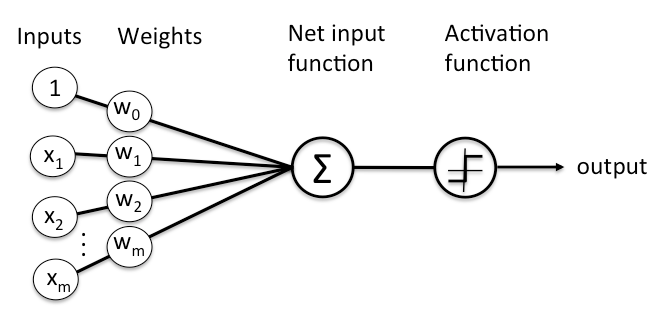}
\vspace{-6ex}
\caption{\label{fig:node}Weighted averaging, followed by
nonlinear activation.}
\vspace{-4ex}
\end{wrapfigure}

\begin{figure}[!h]
\vspace{3ex}
\centering
\includegraphics[width=1\linewidth]{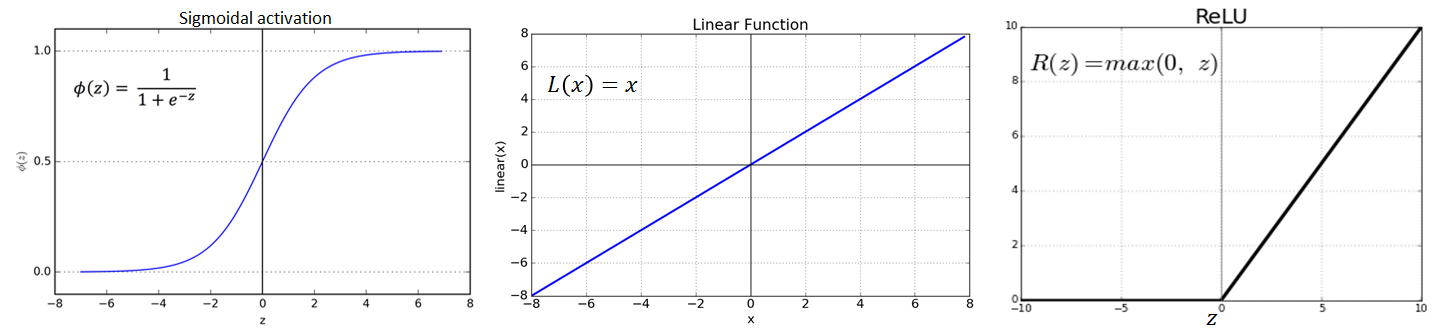}
\caption{Commonly used activation functions in neural networks.}
\label{fig:activations}
\end{figure}

Every node of an NN performs two operations:
weighted averaging of the inputs, and a nonlinear
transformation of the weighted sum using a so-called
\emph{activation function}, see Fig.~\ref{fig:node}.  
Some of the commonly used activation functions are
depicted in Fig.~\ref{fig:activations}.

A neural network is formed by interconnecting several such nodes
using different architectures.  Each connection from node $i$ to
node $j$ is characterized by the weight that is used for the
output of node $i$ for the weighted average performed at node $j$.
Typical architectures consists of layers of nodes connected
to the nodes of the subsequent layer.  The output of the neural
network is vector with each entry representing the output of the
corresponding node in the final layer.  It is common to construct
the network to have $k$ output nodes if the goal is to assign
one (or more) of $k$ possible class assignments.  Moreover,
all the $k$ values lie in $[0, 1]$ and sum to 1.  The input
is assigned the class label $0 \leq l \leq k-1$ if the value
of the $l^{th}$ node is the highest among the $k$ outputs.

\vspace*{-1ex}

We introduce notations for the parameter vector of a neural
network as follows. $NN_p$ denotes
an instance of the neural network with $p$ being the vector
of parameter assignments. Given an input $x$, and the instance
if the neural network $NN_p$, the vector of $k$ outputs is
returned by the function $f_{NN_p}(x)$.  $NN_p(x)$ is the
index of the highest output and thus corresponds to the class
label $l$ assigned to the input.
\vspace{-1ex}

\subsection*{The dReal Solver}
The dReal~\cite{GaoKC13} tool is an SMT solver~\cite{d2008survey}
for nonlinear theories over the reals.The tool can handle first order formula defined by nonlinear real functions such as polynomials, trigonometric functions, exponential functions, etc. It implements the framework of $\delta$-complete decision procedure~\cite{gao2012delta}, which has two possible outputs:
\begin{itemize}
\item \textsf{unsat}: no variable assignment
  satisfying the formula.
\item $\delta$-\textsf{sat}: exists a variable assignment $\xi$ satisfying the formula if we consider a
user-specified numerical perturbation $\delta \in \mathcal{Q}^+$. 
\end{itemize}
We note that the satisfiability of first-order formula over the real is undecidable~\cite{DBLP:conf/hybrid/AlurCHH92}. The tool is
implemented in the framework of delta-complete analysis,
which provides an algorithm for the originally undecidable problem by
using approximation (the use of $\delta$ in the analysis).

The latest version of dReal~\cite{github_dreal} now implements Optimization Modulo Theory (OMT)~\cite{sebastiani2015optimathsat,bjorner2015nuz}. OMT is an extension of SMT which allows for finding models that optimize given objectives.

\section{Parameter Robustness}
\label{sec:param_robust}

In this section, we present various definitions of parameter robustness analysis for neural networks.  

We begin with a definition of parameter robustness locally to an input similar to local input robustness as presented in~\cite{katz_2017,huang2017safety,bastani2016measuring}.

\begin{definition}
An NN with parameter vector $p_0$ is  $(\delta,\varepsilon)$-parameter robust locally at an input $x_0$ if and only if:
\begin{equation}
    \forall p.|p-p_0|\le \delta \implies |f_{NN_{p_0}}(x) - f_{NN_p}(x)|\le \varepsilon 
\end{equation}
\end{definition}

Definition 1 gives a quantitative measure on the change in confidence of labeling a certain input. This definition, however, does not cover all inputs in the input domain. The following definition address this:  

\begin{definition}
An NN with parameter vector $p_0$ is  $(\delta,\varepsilon)$-parameter robust globally for a input domain $\mathcal{D}$ if and only if:
\begin{equation}
    \forall x\in \mathcal{D}, \forall p.|p-p_0|\le \delta \implies |f_{NN_{p_0}}(x) - f_{NN_p}(x)|\le \varepsilon 
\end{equation}
\end{definition}

Though the definitions of parameter robustness described above give a quantitative measure on the change of confidence, it does not say whether the decision label will actually be changed. For example, if the confidence value changes positively for a given label, the decision label will remain the same, even though $\varepsilon$ could be higher. As a result, the above robustness measures 
gives only an idea on relative change in confidence value, but not how the actual label will be changed. 

Now, we define the parameter robustness that specifies whether an actual label of an input changes. Both local and global versions are defined as follows:

\begin{definition}
\label{def:l_flip}
An NN with parameter vector $p_0$ is locally $\delta$-parameter robust locally at an input $x_0$ if and only if:
\begin{equation}
\label{eq:g_flip}
    \forall p.|p-p_0|\le \delta \implies NN_{p_0}(x_0) = NN_p(x_0) 
\end{equation}
\end{definition}

\begin{definition}
\label{def:g_flip}
An NN with parameter vector $p_0$ is locally $\delta$-parameter robust globally for an input domain $\mathcal{D}$ if and only if:
\begin{equation}
\label{eq:g_flip}
    \forall x\in\mathcal{D}, \forall p.|p-p_0|\le \delta \implies NN_{p_0}(x) = NN_p(x) 
\end{equation}
\end{definition}

Definition~\ref{def:g_flip} states that for an NN to be $\delta$-parameter robust globally for all input in the domain, no input cannot be mislabeled. This is rather a very strict definition of robustness. In particular, when a quantization technique is applied to NN, it is expected that the labels for some inputs will be changed, at least the inputs close to the decision boundary. To incorporate this, we slightly modify the definition~\ref{def:g_flip} as follows:

\begin{definition}
An NN with parameter vector $p_0$ is locally $\delta_\sigma$-parameter robust globally for an input domain $\mathcal{D}$ if and only if:
\begin{equation}
\label{eq:g_flip}
    \forall x\in\mathcal{D}, \forall p.|p-p_0|\wedge |f_{NN_{p_0}}(x)-l|\ge \sigma \le \delta \implies NN_{p_0}(x) = NN_p(x) 
\end{equation}
where $l$ denotes the level set of the confidence function, which is used to label the input, i.e, $f_{NN_{p_0}}(x)=l$ represents a decision boundary.
\end{definition}

\begin{figure}
    \centering
    \includegraphics[scale=0.45]{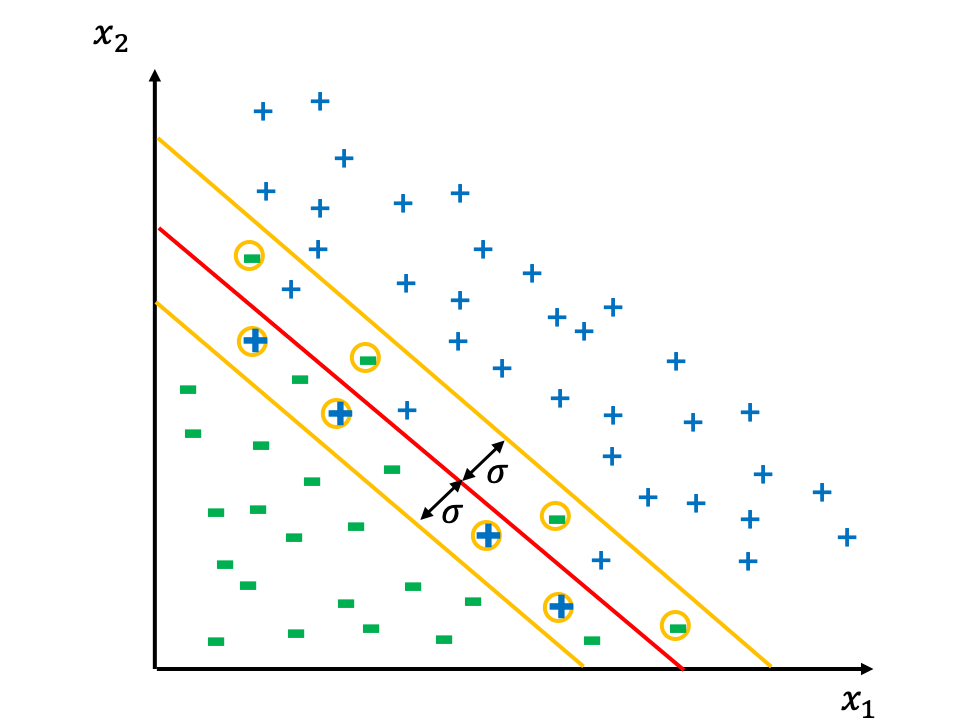}
    \vspace{-2ex}
    \caption{Illustration of $\delta_\sigma$-parameter robustness on a two-class classifier.}
    \label{fig:del_sig_robust}
\vspace*{-8ex}
\end{figure}

The $\delta_\sigma$-parameter robustness of NN is illustrated in~\ref{fig:del_sig_robust}. The red line represents the decision boundary and `-' and `+' represent the decision labels. The yellow lines are $\sigma$ distance away from the decision boundary. Definition 5 states that all the inputs that are $\sigma$ or more distance away from the decision boundary (i.e., all the points either above the top yellow line or below the bottom yellow line) will be labeled as same in both $NN_{p_0}$ and $NN_p$. The inputs between the yellow lines, however, may be be mislabeled, as illustrated by the points inside the yellow circles in the figure.
\vspace{-2ex}

\section{Verification and Estimation of Parameter Robustness}
\label{sec:ver}
In this section we will present how to verify and estimate parameter robustness using SMT solver. 

\subsection{Verifying Parameter Robustness}
We apply SMT solver to verify all the parameter robustness defined in Section~3. The key idea is to construct a formula for each of them by the negating their definition. The robustness property will then be verified if the SMT solver returns \textsf{unsat}. The formula for all the parameter robustness given to SMT solver are as follows:

\begin{itemize}
\item To verify Eq.~1, we use the following formula:
    \begin{equation}
        \exists p. (p\ge p_0-\delta)\wedge (p\le p_0+\delta)\wedge abs(f_{NN_{p_0}}(x_0 )-f_{NN_p}(x_0))>\varepsilon 
    \end{equation}
\item To verify Eq.~2, we use the following formula:
    \begin{equation}
    \begin{split}
        &\exists p,x, (p\ge p_0-\delta)\wedge (p\le p_0+\delta)\wedge (x\ge \underline{x})\wedge (x\le \bar{x})
        \wedge abs(f_{NN_{p_0}}(x_0 )-\\ 
        & f_{NN_p}(x_0))>\varepsilon 
        \end{split}
    \end{equation}
where we define the input domain $\mathcal{D}$ as a bounding box, i.e., $\mathcal{D}=[\underline{x},\bar{x}]$
\item To verify Eq.~3, we use the following formula:
    \begin{equation}
    \begin{split}
        &\exists p. (p\ge p_0-\delta)\wedge (p\le p_0+\delta)\wedge ((f_{NN_{p_0}}(x_0)\le l)\wedge 
        (f_{NN_p}(x_0) > l)\vee \\
        &(f_{NN_{p_0}}(x_0)< l)\wedge (f_{NN_p}(x_0) \ge l))
    \end{split}    
    \end{equation}
    Here we encode $NN_{p_0}(x_0)=NN_p(x_0)$ as follows:
    \begin{equation*}
      ((f_{NN_{p_0}}(x_0)\le l)\wedge 
        (f_{NN_p}(x_0) \le l))\vee \\
        ((f_{NN_{p_0}}(x_0) > l)\wedge (f_{NN_p}(x_0) > l)) 
    \end{equation*}
    That is $x_0$ falls in the same side of the decision boundary both in $NN_{p_0}$ and $NN_p$. For the verification purpose, we consider its negation.
\item To verify Eq.~4, we use the following formula:
    \begin{equation}
    \begin{split}
        &\exists p, x. (p\ge p_0-\delta)\wedge (p\le p_0+\delta)
        \wedge (x\ge \underline{x})\wedge (x\le \bar{x})
        \wedge ((f_{NN_{p_0}}(x_0)\le l)\\
        &\wedge (f_{NN_p}(x_0) > l)
        \vee(f_{NN_{p_0}}(x_0)< l)\wedge (f_{NN_p}(x_0) \ge l))
    \end{split}    
    \end{equation}
\item To verify Eq.~5, we use the following formula:
    \begin{equation}
    \begin{split}
        &\exists p, x. (p\ge p_0-\delta)\wedge (p\le p_0+\delta)
        \wedge (x\ge \underline{x})\wedge (x\le \bar{x}) \wedge (abs(f_{NN_{p_0}}(x)-l)\ge \sigma) \\
        &\wedge ((f_{NN_{p_0}}(x_0)\le l)
        \wedge (f_{NN_p}(x_0) > l))
        \vee((f_{NN_{p_0}}(x_0)< l)\wedge (f_{NN_p}(x_0) \ge l))
    \end{split}    
    \end{equation}
\end{itemize}

We verify all the robustness properties on dReal solver~\cite{github_dreal}.

\subsection{Estimating Maximum Parameter Robustness}
For $(\delta,\varepsilon)$-parameter robustness, we allow $\delta$-perturbation on the parameter space and check whether the confidence value is  bounded by $\varepsilon$. The estimation problem is defined as computing maximum possible value of $\varepsilon$ for a given value of $\delta$. We are interested in this estimation problem, as the maximum value of $\varepsilon$ represents the least robustness measure for a given $\delta$ value. The estimation problem can be formulated as an optimization problem as follows:
\begin{itemize}
\item $\varepsilon$-Estimation for $(\delta, \epsilon)$-parameter robustness locally at $x_0$:
\begin{equation}
\begin{aligned}
&\underset{\varepsilon\in[0,\bar{\varepsilon}]}{\text{minimize }}
-\varepsilon \\
&\text{subject to:}\\
& (p\ge p_0-\delta)\\ 
& (p\le p_0+\delta) \\
& abs(f_{NN_{p_0}}(x_0 )-f_{NN_p}(x_0))=\varepsilon 
\end{aligned}
\end{equation}

where, $\bar{\varepsilon}$ is the maximum value of $\varepsilon$. Note that, instead of maximizing, $\varepsilon$, we minimize its negation, as the SMT solver we used implements only the minimization problem. 
\item $\varepsilon$-Estimation for $(\delta, \epsilon)$-parameter robustness globally for $\mathcal{D}=[\underline{x},\bar{x}]$:
\vspace*{-1ex}
\begin{equation}
\begin{aligned}
&\underset{\varepsilon\in[0,\bar{\varepsilon}]}{\text{minimize }}
-\varepsilon \\
&\text{subject to:}\\
& (p\ge p_0-\delta)\\ 
& (p\le p_0+\delta) \\
& (x \ge \underline{x})\\
& (x \le \bar{x})\\
& abs(f_{NN_{p_0}}(x_0 )-f_{NN_p}(x_0))=\varepsilon 
\end{aligned}
\end{equation}
\end{itemize}
Similarly, for $\delta_\sigma$-parameter robustness, we consider estimation problem for $\sigma$. For a given value $\delta$, we want to maximize $\sigma$, which tells us how far away the boundary needs to be shifted so that no input beyond it cannot be mislabeled. We formulate this estimation problem as follows:
\begin{equation}
\begin{aligned}
&\underset{\sigma\in[0,\bar{\sigma}]}{\text{minimize }}
-\sigma \\
&\text{subject to}\\
& (p\ge p_0-\delta)\\ 
& (p\le p_0+\delta) \\
& (x \ge \underline{x})\\
& (x \le \bar{x})\\
&  abs(f_{NN_{p_0}}(x)- l))=\sigma\\
&(f_{NN_{p_0}}(x)\le l)\wedge 
        ((f_{NN_p}(x) \le l))\vee 
        ((f_{NN_{p_0}}(x) > l)\wedge (f_{NN_p}(x) > l)) 
\end{aligned}
\end{equation}
where maximum value of $\bar{\sigma}$ is the maximum value of $\sigma$.

\section{Case Studies}
\label{sec:case_studies}
We describe two datasets and the corresponding neural networks as
case studies for our robustness analysis framework.
\begin{figure}[!h]
\subfigure[ReLU activation.]
{\includegraphics[width=0.49\linewidth]{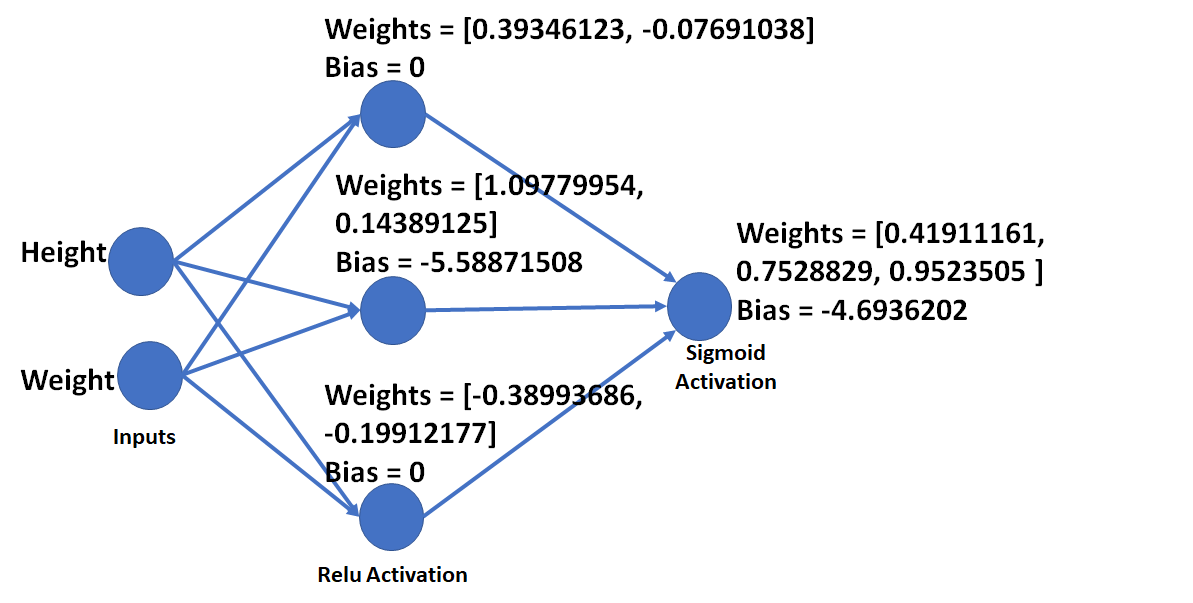}\label{fig:athletes_relu}}
\subfigure[Linear activation.]
{\includegraphics[width=0.49\linewidth]{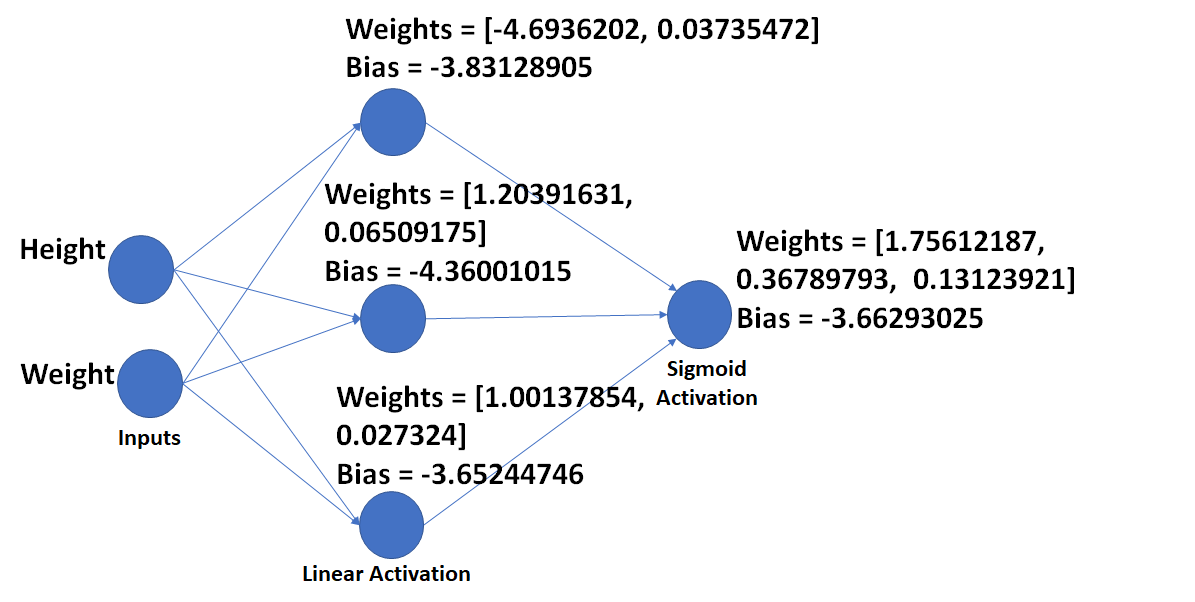}\label{fig:athletes_linear}
}
\caption{Two MLPs trained on the Athletes dataset.}
\label{fig:athletes}
\vspace{-3ex}
\end{figure}

The first dataset, known as \emph{cats}, contains the height,
weight, and gender of 144 domesticated cats (47 female and 97 male)
\cite{cats_dataset}.  The gender identification problem entails
learning a classifier to estimate if a cat is Male or Female based
on its height and weight.  We present a simple one-node model that
implements logistic regression and examine its robustness.

Given the height ($H$) and the weight ($W$) of a cat, the classifier,
learned using Python Scikit-Learn 0.20.3, is given by $y = sig(x)$,
where $x = c_0 + c_1H + c_2W$.  We assign the class label ``Male'' if
the $y \geq 0.5$ and ``Female'' otherwise.  The parameters of the
model that were learned on 78\% of the data were $c_0 = -3.51518067,
c_1 = 0.07577862$, and $c_2 = 1.18118408$.  The multinomial loss
function was optimized using the \emph{lbgfs} algorithm
\cite{scikit-learn}.  The testing accuracy on the remaining 22\%
was 87.5\%.

A second dataset contains the official statistics on the 11, 538
athletes (6,333 men and 5,205) men that participated in the 2016
Olympic games at Rio de Janeiro~\cite{athletes_dataset}.  Each row
contains an id, the name, nationality, gender, date birth, height,
weight, sport of the athletes and the medals tally.  The gender
identification problem entails learning an MLP to guess the gender
of the athlete based on their height and weight.  We present two
MLPs for this problem and examine their robustness in the next section.

Given the height and weight of an athlete as the input, the MLPs
is constructed using two layers: a hidden layer and an output node.
Three nodes that make up the hidden layer perform weighted averaging
of the inputs and transform them using a nonlinear activation.  Their
outputs are then fed to the output node, which again takes a weighted
average and uses the Sigmoid activation to obtain a number between 0
and 1.  If the output is greater than 0.5, the input is assigned 
``Male'', otherwise it is assigned ``Female''.  We implemented two
variations of the model.  We used ReLU and linear activations in the
three nodes of the hidden layer.  The two models and their parameters
are illustrated in Fig.~\ref{fig:athletes}.

The models were implemented in Keras and trained on a GPU-based
instance of Amazon Web Services.  The training accuracy was 77.19\%
and 77.36\% for the ReLU and Linear versions respectively after
200 epochs.

In the next section, we apply our robustness analysis framework
to examine the effect of quantizing the parameters of the
logistic regression model for the Cats dataset and the two MLPs
for the athletes dataset.
\section{Results}
\label{sec:results}
In this section we discuss our results. For all three NNs (one for first case study and two for second case study), we present results for $(\delta,\varepsilon)$- and $\delta_\sigma$-parameter robustness both locally at an input and globally at input domain.
\begin{table}
\begin{center}
\bgroup
\def\arraystretch{1.5}%
\vspace{-2ex}
\begin{tabular}{|c|c|c|c|}
\hline 
\multirow{2}{*}{$\delta$} &	
      \multicolumn{3}{c|}{$\varepsilon$}
      \\
      \cline{2-4}
     & CAT &  ATH-ReLU & ATH-Linear \\
	\hline 
	0.005 & 0.00691 & 0.166 & 0.545\\
	\hline
	0.01 & 0.05054 & 0.0825 & 0.219\\
    \hline
    \multirow{2}{*}{$\delta$} &	
      \multicolumn{3}{c|}{$\sigma$}
      \\
      \cline{2-4}
     & CAT &  ATH-ReLU & ATH-Linear \\
     \hline
     0.005 & (0.024, 0.021) & (0.082, 0.076) & (0.268, 0.218)\\
     \hline
     0.01 & (0.052, 0.04) & (0.165, 0.144) & (0.44, 0.34) \\
	\hline
\end{tabular}
\egroup
\end{center}
\label{table:res}
\caption{Estimated values for $\varepsilon$ and $\sigma$ for $(\delta,\varepsilon)$- and $\delta_\sigma$-parameter robustness, respectively,
globally for input domains}
\vspace*{-4ex}
\end{table}

Table~1 shows the estimated value of $\varepsilon$ of $(\delta,\varepsilon)$-parameter robustness, computed using Eq.~(12), and $\sigma$ of $\delta_\sigma$-parameter robustness, computed using Eq.~(13), for entire input domains. We compute them for two different $\delta$ values. The column CAT, ATH-ReLU and ATH-Linear represent the results for cat classifier, athletic classifier with ReLU activation and athletic classifier with Linear activation, respectively. The tuple in the table for $\sigma$  represents values for  \textit{male} and \textit{female} class, repectively.   If we compare the results of ATH-ReLU and ATH-Linear, it is clear that the former classifier is much more robust than the latter for $\delta$ pertubation of the parameter values.

\vspace{-0ex}
\begin{figure}[!h]
\subfigure[\scriptsize $(\delta,\varepsilon)$-parameter robustness for $\delta=0.005$]
{\includegraphics[width=0.49\linewidth]{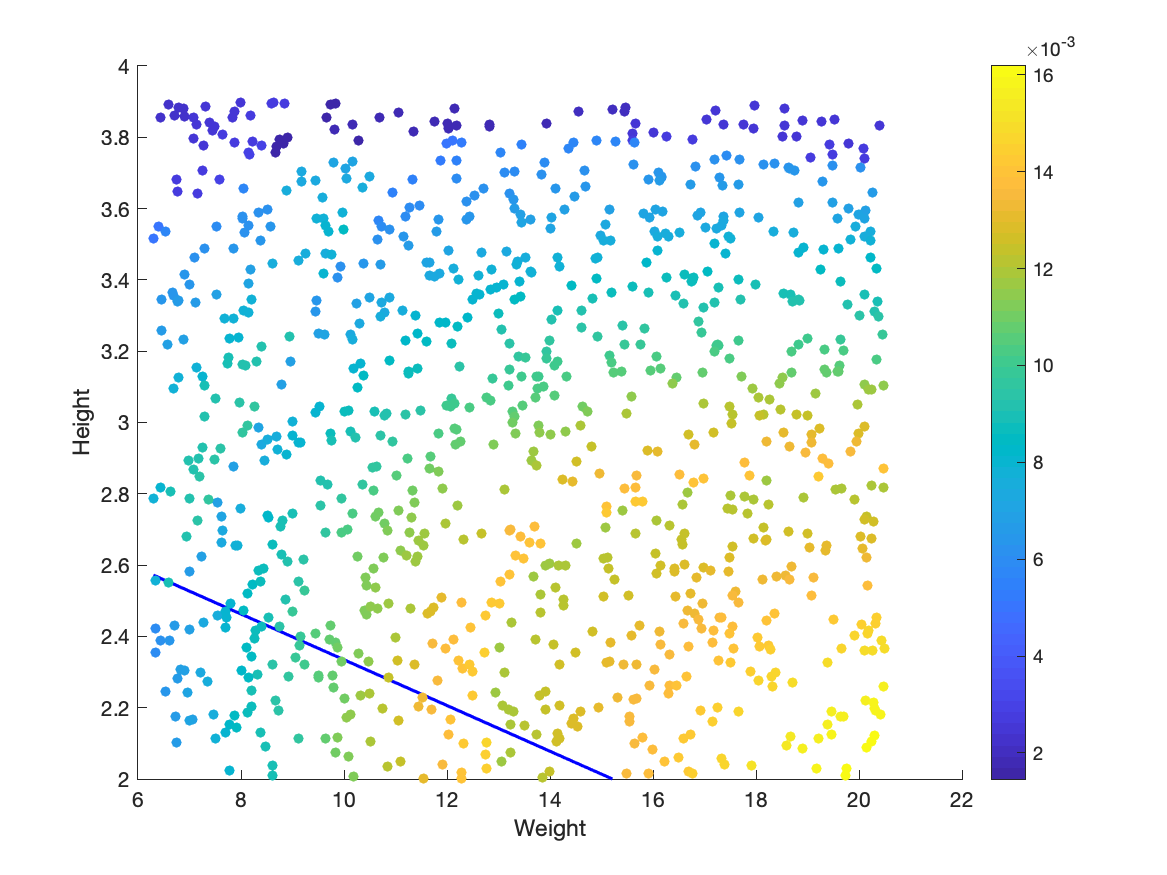}}
\subfigure[\scriptsize $(\delta,\varepsilon)$-parameter robustness for $\delta=0.01$]
{\includegraphics[width=0.49\linewidth]{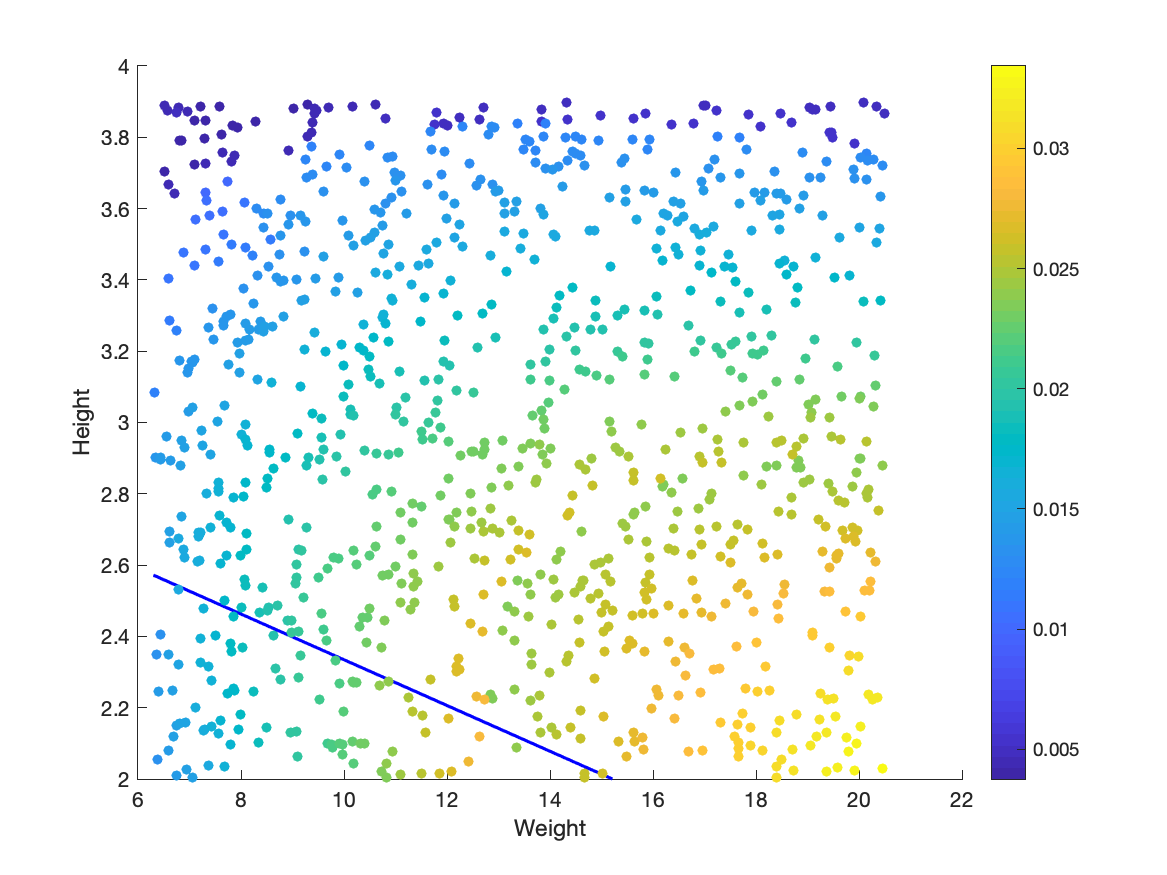}}
\subfigure[\scriptsize $\delta_\sigma$-parameter robustness for $\delta=0.005$]
{\includegraphics[width=0.49\linewidth]{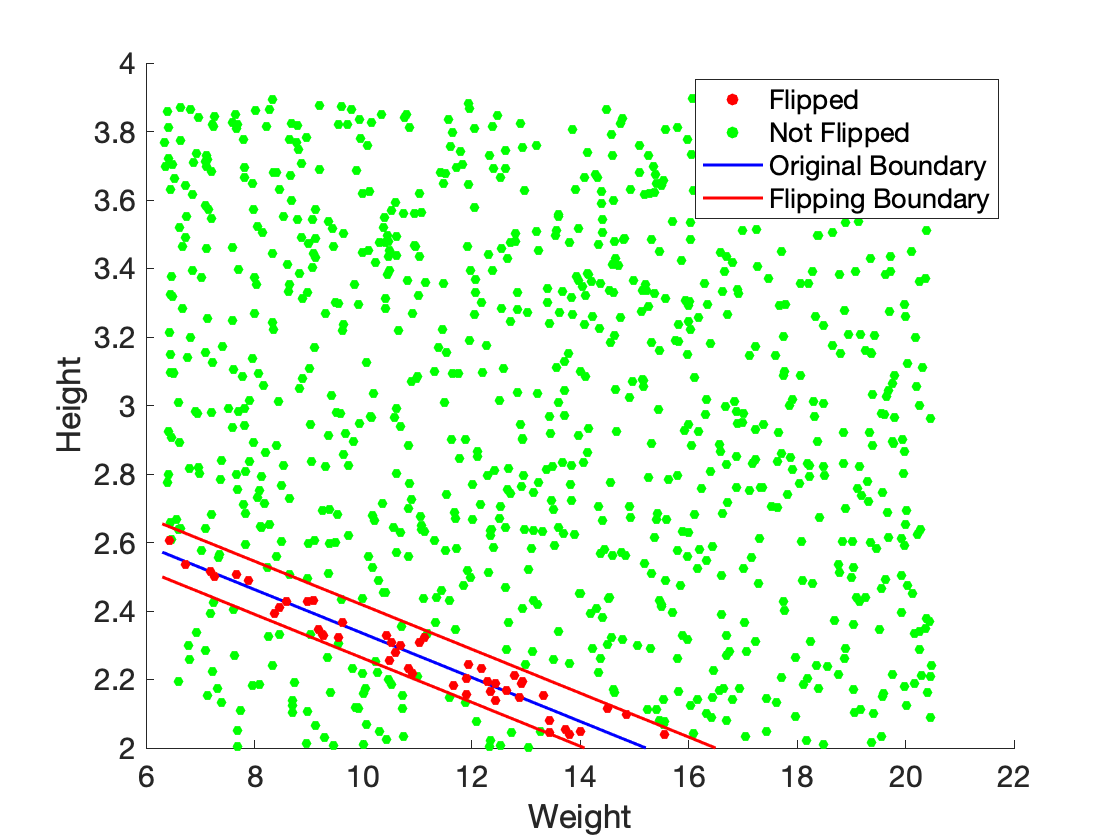}}
\subfigure[\scriptsize $\delta_\sigma$-parameter robustness for $\delta=0.01$]
{\includegraphics[width=0.49\linewidth]{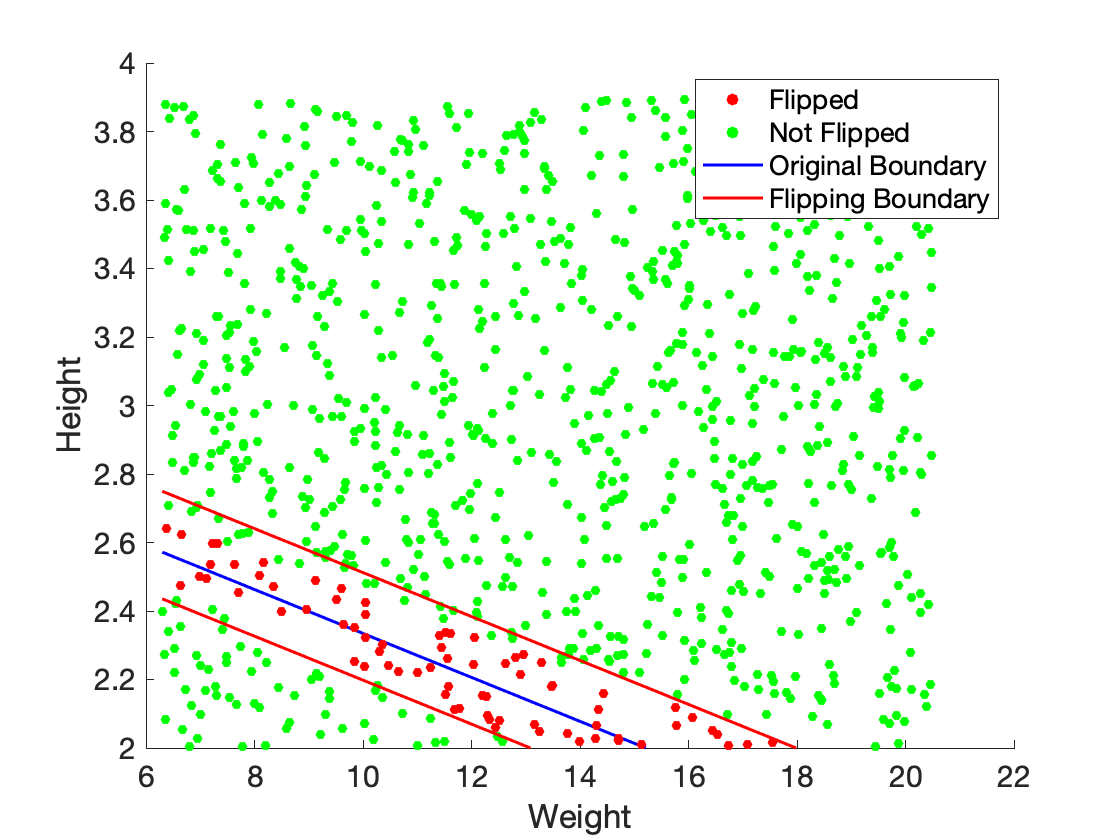}}
\vspace{-2ex}
\caption{Parameter robustness analysis of Cat classifier.}
\label{fig:res_cat}
\vspace*{-5ex}
\end{figure}

Fig.~\ref{fig:res_cat} illustrates parameter robustness of the Cat classifier presented. For $(\delta,\varepsilon)$-parameter robustness, we choose two different $\delta$ values ($0.005$ and $0.01$). For both cases, we randomly chose $1000$ points from the input domain. We then computed $\varepsilon$ for all inputs using Eq.~(11). Fig.~\ref{fig:res_cat}(a,b) show $(\delta,\varepsilon)$-parameter robustness locally at each randomly selected points. The blue line represents the decision boundary of NN, whereas the colorbar represents the range of $\varepsilon$. It is clear from the figures that $\varepsilon$ value is higher in the bottom right region, which means the region is more susceptible to be mislabeled in the perturbed network. Note that does not mean that the input would actually be mislabeled (see explanation in Section~3).  

Fig.~6(c,d) illustrate both $\delta$- and $\delta_\sigma$-parameter robustness for two different $\delta$ values. For $\delta$-parameter robustness, we selected $1000$ random inputs from the domain. We then checked whether the input labeled will be flipped in the perturbed network using Eq.~(8). In the figures, green and red points represent \textit{non-flippable} and \textit{flippable} inputs, respectively. The top (bottom) red line is generated by adding (subtracting) $\sigma$ to the decision boundary, where $\sigma$ is computed using Eq.~(13). 
\vspace{-0ex}
\begin{figure}[!h]
\subfigure[\scriptsize $(\delta,\varepsilon)$-parameter robustness for $\delta=0.005$]
{\includegraphics[width=0.49\linewidth]{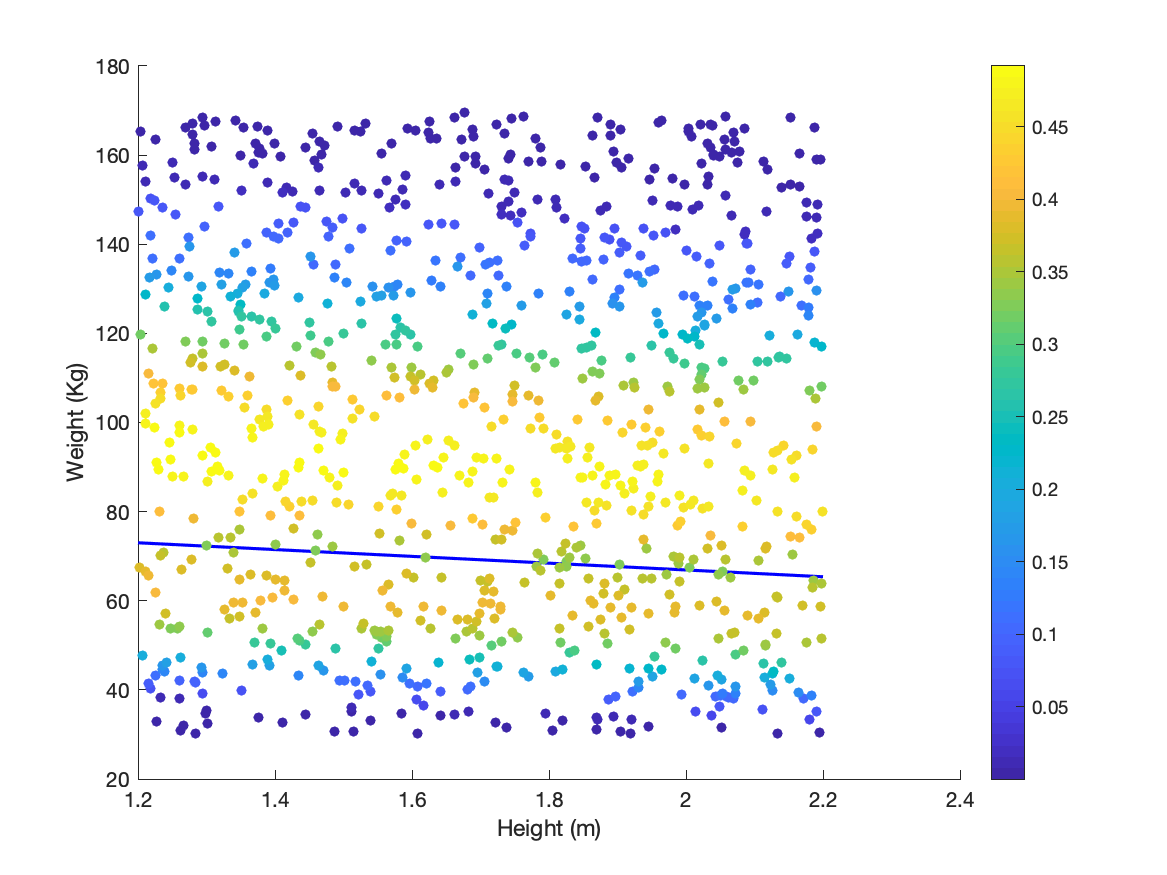}}
\subfigure[\scriptsize $(\delta,\varepsilon)$-parameter robustness for $\delta=0.01$]
{\includegraphics[width=0.49\linewidth]{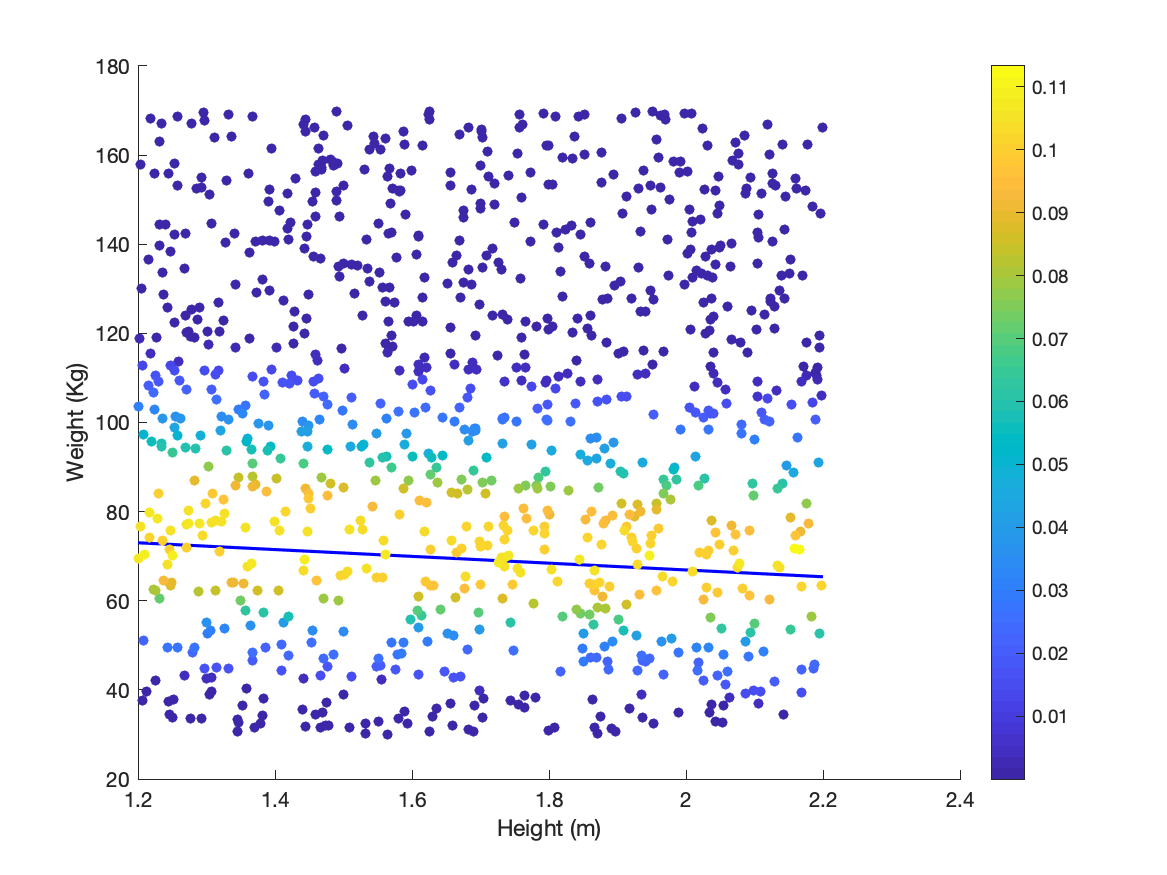}}
\subfigure[\scriptsize $\delta_\sigma$-parameter robustness for $\delta=0.005$]
{\includegraphics[width=0.49\linewidth]{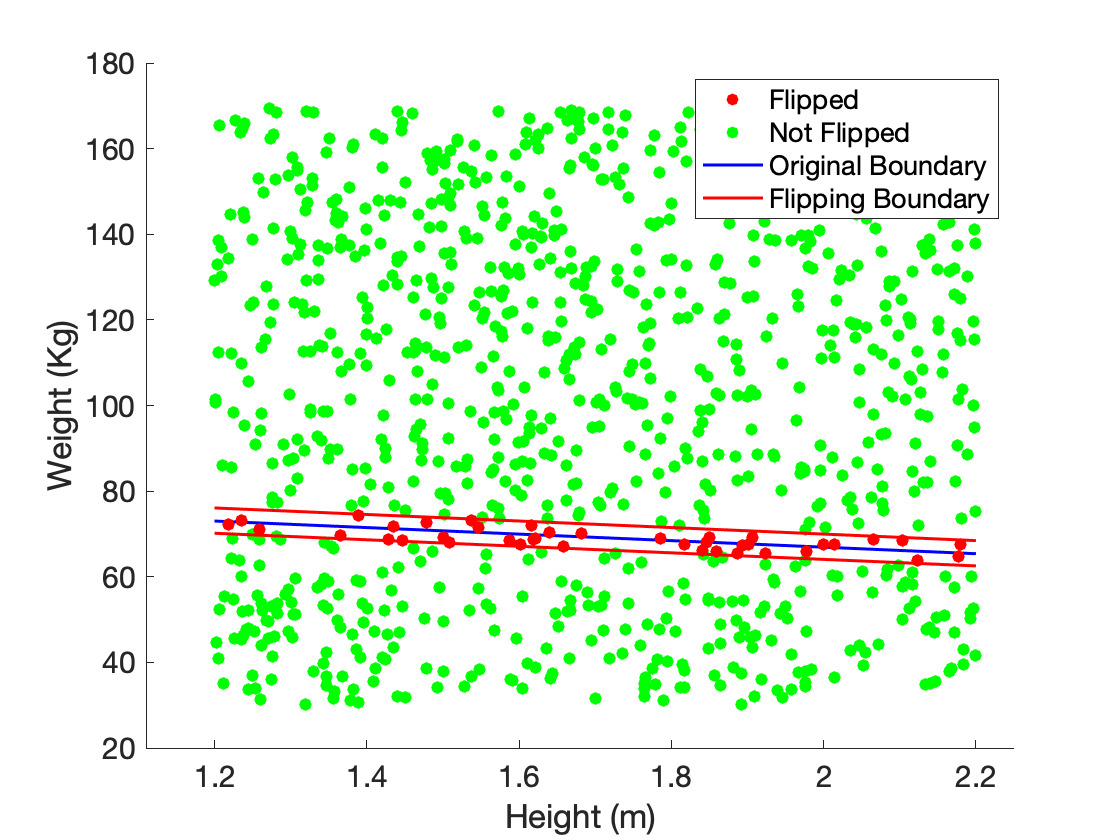}}
\subfigure[\scriptsize $\delta_\sigma$-parameter robustness for $\delta=0.01$]
{\includegraphics[width=0.49\linewidth]{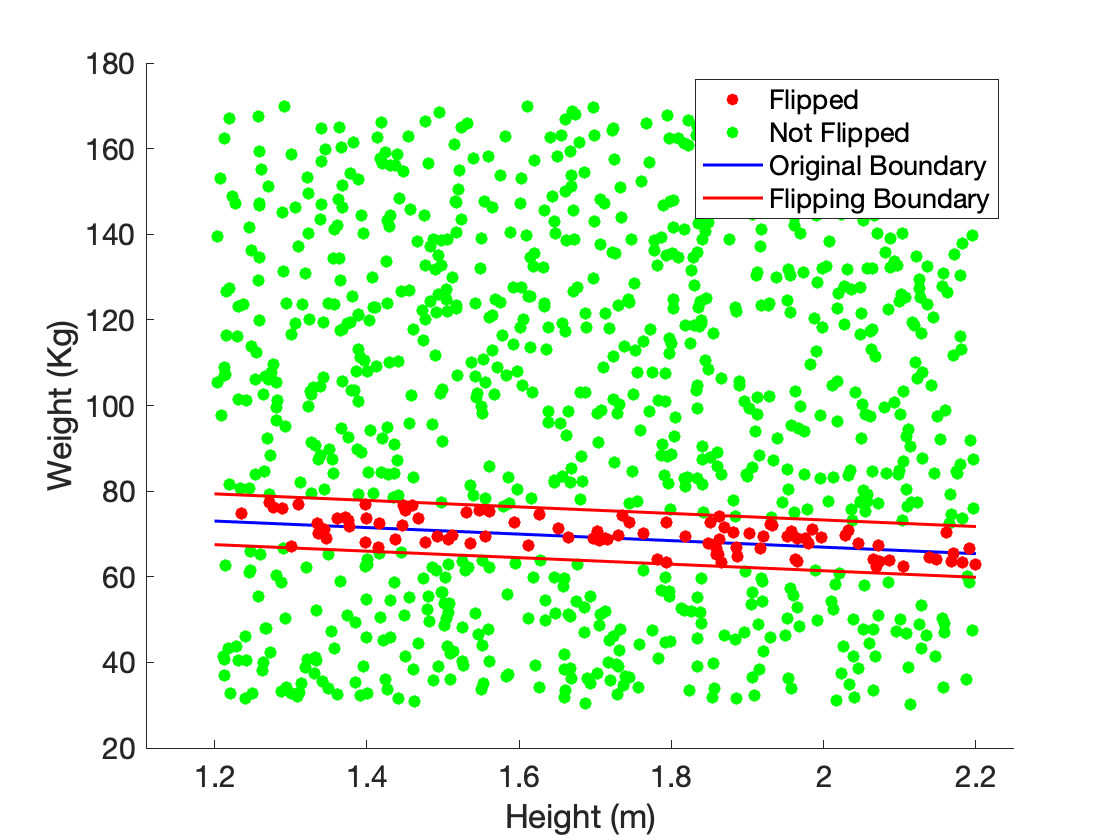}}
\caption{Parameter robustness analysis of Athletics classifier with ReLU activation.}
\label{fig:res_relu}
\vspace{-5ex}
\end{figure}

\begin{figure}[!h]
\subfigure[\scriptsize $(\delta,\varepsilon)$-parameter robustness for $\delta=0.005$]
{\includegraphics[width=0.49\linewidth]{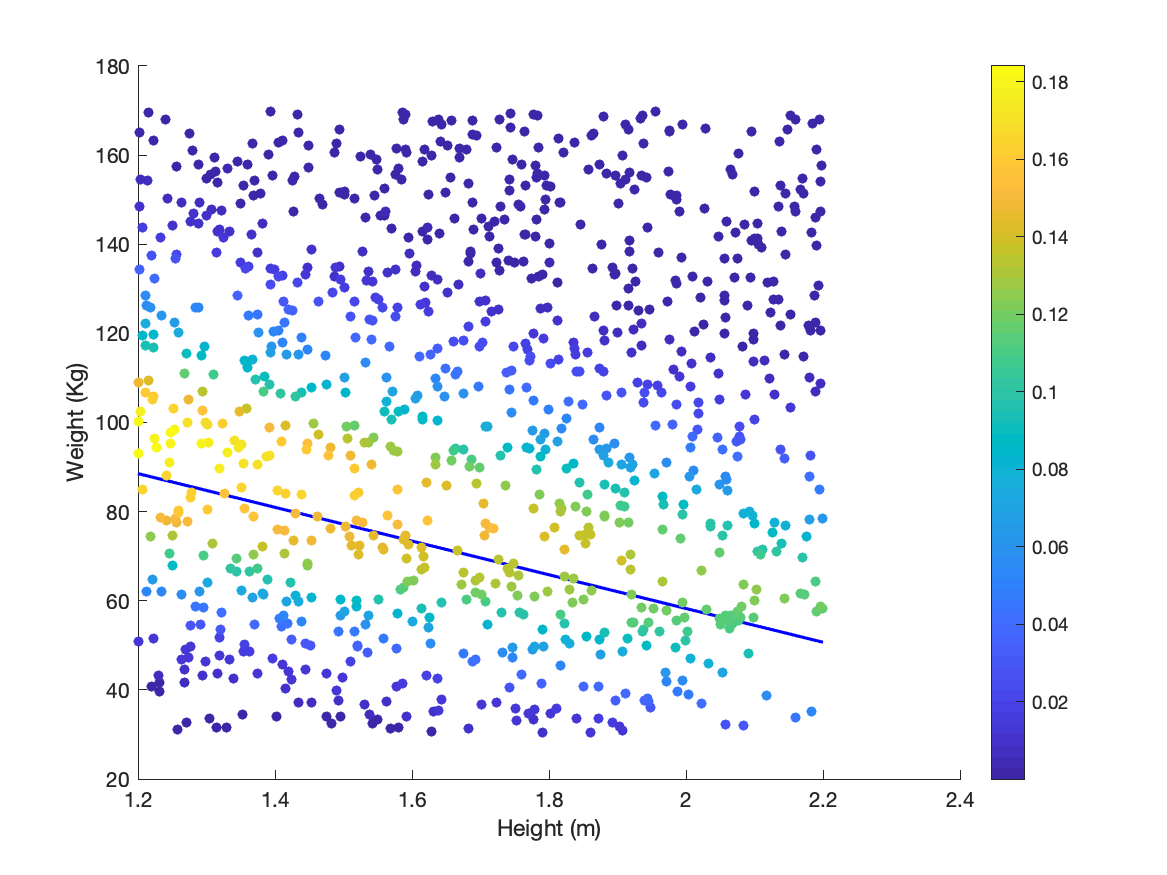}}
\subfigure[\scriptsize $(\delta,\varepsilon)$-parameter robustness for $\delta=0.01$]
{\includegraphics[width=0.49\linewidth]{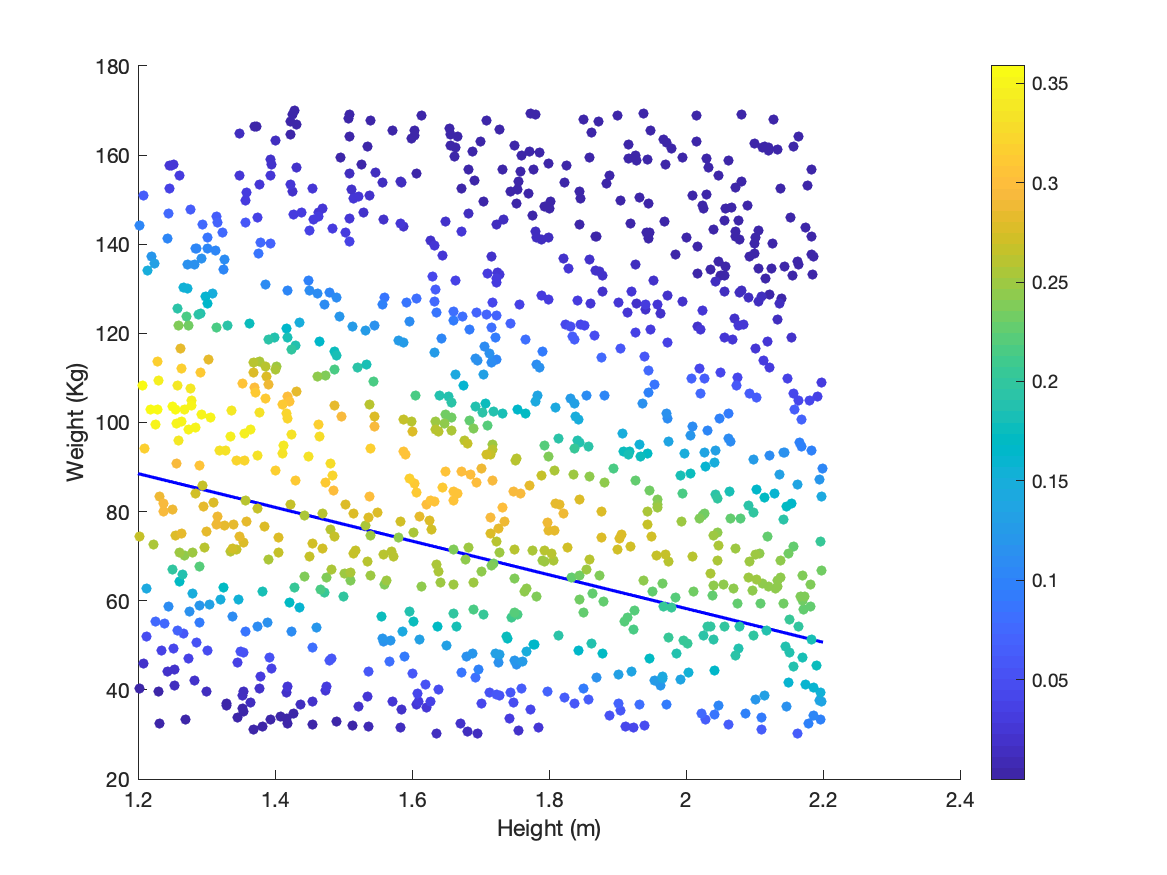}}
\subfigure[\scriptsize $\delta_\sigma$-parameter robustness for $\delta=0.005$]
{\includegraphics[width=0.49\linewidth]{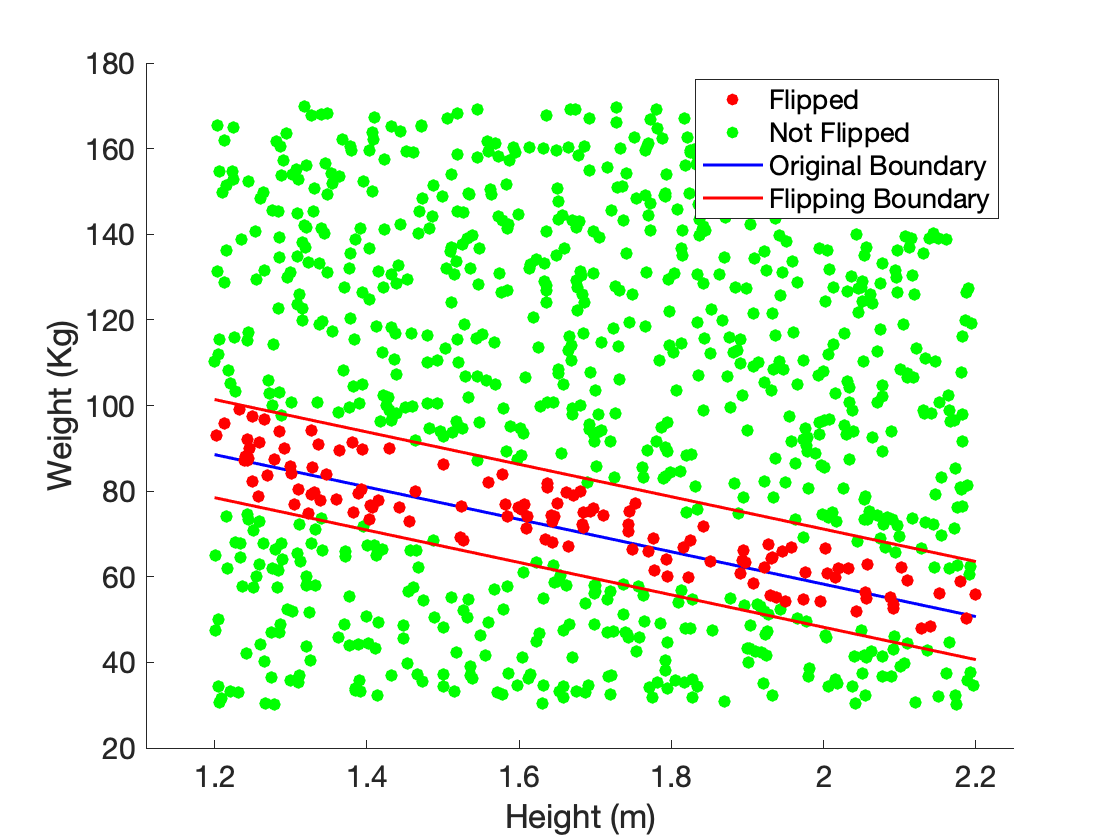}}
\subfigure[\scriptsize $\delta_\sigma$-parameter robustness for $\delta=0.01$]
{\includegraphics[width=0.49\linewidth]{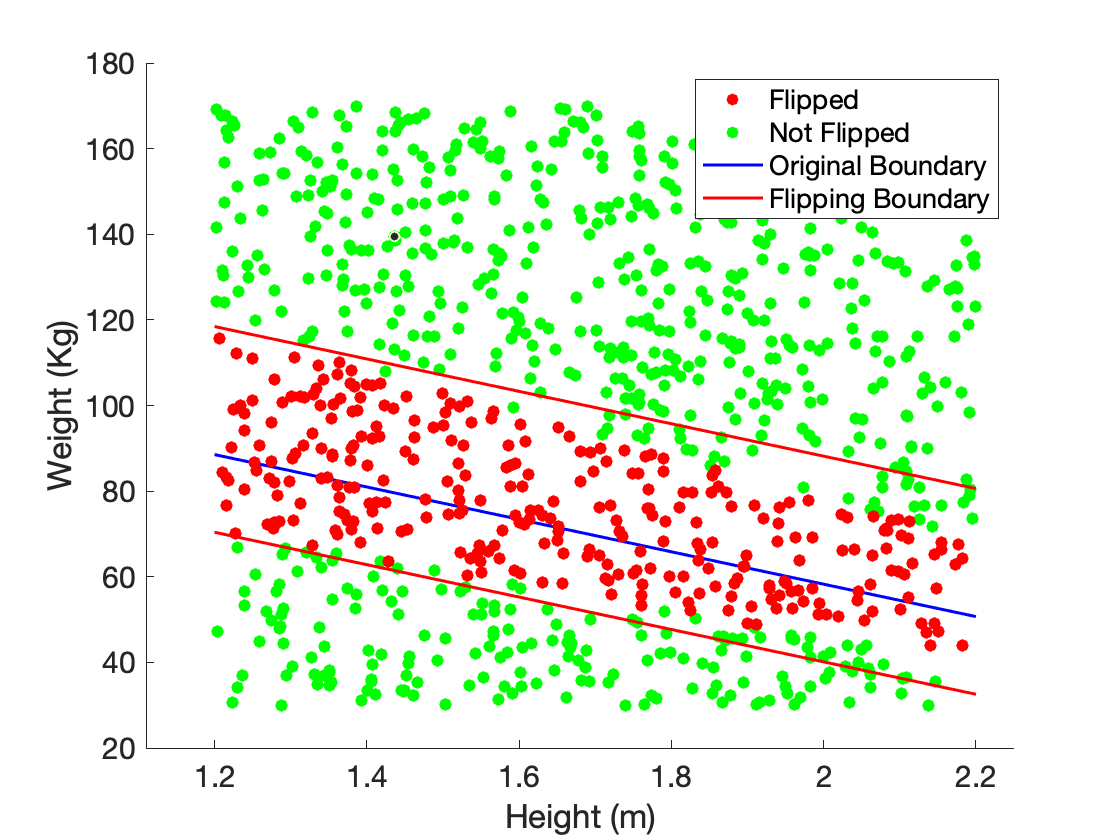}}
\vspace{-1ex}
\caption{Parameter robustness analysis of Athletes classifier with linear activation.}
\label{fig:res_lin}
\vspace{-5ex}
\end{figure}

Figs.~\ref{fig:res_relu} and \ref{fig:res_lin} illustrate the parameter robustness analysis of the athletics classifier with ReLU and Linear activation, respectively. Comparing these figures, we can conclude that
the athletics classifier with ReLU activation is much more robust as compared to the classifier with linear activation.
\vspace{-2ex}

%

\section{Related Work}
\label{sec:related}
\vspace{-2ex}
Robustness analysis of neural networks is an active area of
research.  In this section, we compare and contrast some of
the recent papers with our framework.  Robustness typically refers
to an NN's ability to handle perturbations in the input data.
The efforts to characterize robustness can be broadly classified
into two types: model-centric approaches and data-centric approaches.

Model-centric approaches focus on improving the problem formulation
to construct robust networks.  Distillation training, one of the
earliest attempts, entails training one model to predict the output
probabilities of another model that was trained on an earlier,
baseline standard to emphasize accuracy
\cite{distillation,distillation-dnn}.  In \cite{carlini_2016}, the
authors proposed a new set of attacks for the $L_{0}$, $L_{2}$, and
$L_{\infty}$ distance metrics to construct upper bounds on the
robustness of neural networks and thereby demonstrate that defensive
distillation is limited in handling adversarial examples.
Adversarial perturbations, random noise, and geometric
transformations were studied in \cite{fawzi_2017} and the authors
highlight close connections between the robustness to additive
perturbations and geometric properties of the classifier’s decision
boundary, such as the curvature.  Spatial Transform Networks, which
entail geometrical transformation of the a network's filter maps were
proposed in \cite{spatial} to improve the robustness to
geometric perturbations.  
Recently, a generic analysis framework CROWN was proposed to certify
NNs using linear or quadratic upper and lower bounds for general
activation functions~\cite{zhang_2018}.  The authors extended their work
to overcome the limitation of simple fully-connected layers and ReLU
activations to propose CNN-Cert.  The new framework can handle various
architectures including convolutional layers, max-pooling layers, batch
normalization layer, residual blocks, as well as general activation functions
and capable of certifying robustness on general convolutional neural networks
\cite{boopathy_2019}. 

Data-centric approaches entail identifying and rejecting perturbed samples,
or increasing the training data to handle perturbations appropriately.
Binary detector networks that can spot adversarial samples
\cite{binary-detector,safetyNet}, and augmenting data to reflect different
lighting conditions~\cite{aipr_2018} are typical examples.
Additionally, robust
optimization using saddle point(min-max) formulation \cite{robust-optimization}
and region-based classification by assembling information in a hypercube
centered \cite{region-based} have also shown promising results. 
The above-mentioned approaches focus on perturbations to data, but our
framework focuses on perturbations to the parameters with the end
goal of safely implementing the neural networks on resource-constrined platforms.

\section{Conclusions and Directions for Future Work}
\label{sec:conclusions}
\vspace{-1ex}
We presented a framework to automatically estimate the impact
of rounding-off errors in the parameters of a neural network.
The framework uses SMT solvers to estimate the local and global
robustness of a given network.  We applied our framework on a
single-node logistic regression model and two small MLPs.
We will consider convolutional neural networks in the future
and investigate the scalability of our framework to larger
parameter vectors.  Compositionality will be critical to
analyzing real-world neural networks and we will explore
extending the theory of approximate bisimulation and
the related Lyapunov-like functions to our problem.
\vspace{-2ex}
%
%
%
\bibliographystyle{splncs04}
\bibliography{smolkafest}
\end{document}